\documentclass[letterpaper, 10 pt, conference]{ieeeconf}  %
\IEEEoverridecommandlockouts                              %
\usepackage[noadjust,sort]{cite}
\usepackage{amsmath,amssymb,amsfonts,graphicx,textcomp,xcolor,multirow,subcaption,tabu,booktabs}
\usepackage{algorithmic}
\usepackage[ruled,vlined]{algorithm2e}
\usepackage[normalem]{ulem}
\usepackage[pagebackref=true,breaklinks=true,colorlinks=true,bookmarks=false,citecolor=blue]{hyperref}
\hypersetup{
colorlinks=true,
linkcolor=blue,
filecolor=magenta,      
urlcolor=blue,
citecolor=blue
}

\useunder{\uline}{\ul}{}
\title{\LARGE \bf Generalization in Dexterous Manipulation via\\Geometry-Aware Multi-Task Learning}
\author{Wenlong Huang$^{1}$, Igor Mordatch$^{2}$, Pieter Abbeel$^{1}$, Deepak Pathak$^{3}$\\
$^{1}$UC Berkeley, $^{2}$Google Brain, $^{3}$Carnegie Mellon University}
\begin{document}
\maketitle
\thispagestyle{empty}
\pagestyle{empty}
\begin{abstract}
Dexterous manipulation of arbitrary objects, a fundamental daily task for humans, has been a grand challenge for autonomous robotic systems. Although data-driven approaches using reinforcement learning can develop \textit{specialist} policies that discover behaviors to control a single object, they often exhibit poor generalization to unseen ones. In this work, we show that policies learned by existing reinforcement learning algorithms can in fact be \textit{generalist} when combined with multi-task learning and a well-chosen object representation. We show that a single generalist policy can perform in-hand manipulation of over 100 geometrically-diverse real-world objects and generalize to new objects with unseen shape or size. Interestingly, we find that multi-task learning with object point cloud representations not only generalizes better but even outperforms the single-object specialist policies on both training as well as held-out test objects. Video results at~\url{https://huangwl18.github.io/geometry-dex}.
\end{abstract}

\section{Introduction}
Hand dexterity is fundamental to daily human activities, requiring complex and precise control of finger movements. While most animals exhibit fine-grained motor controls for movement and many show elementary skills for manipulation, the functionality and the efficacy of human dexterity is unrivaled by others, rendering humans the unique ability to perform most sophisticated tasks with ease. Therefore, developing similarly-intelligent control strategies for robotic hands has been considered a holy grail for research on autonomous robots. However, it is an exceptionally challenging problem because of its high-dimensional action space and complex intermittent hand-object interactions.

Reinforcement learning (RL) has been growing in popularity for solving sensorimotor control tasks because of its capability to discover behaviors without laborious manual engineering~\cite{heess2017emergence,levine2016end}. Prior works have also demonstrated that with large-scale compute, RL policies can acquire useful skills to control dexterous hands from purely simulated experience and perform complex tasks in the real world, including in-hand rotation~\cite{andrychowicz2020learning} and solving Rubik's Cube~\cite{akkaya2019solving}.

\begin{figure}[t]
 \centering
 \includegraphics[scale=.25]{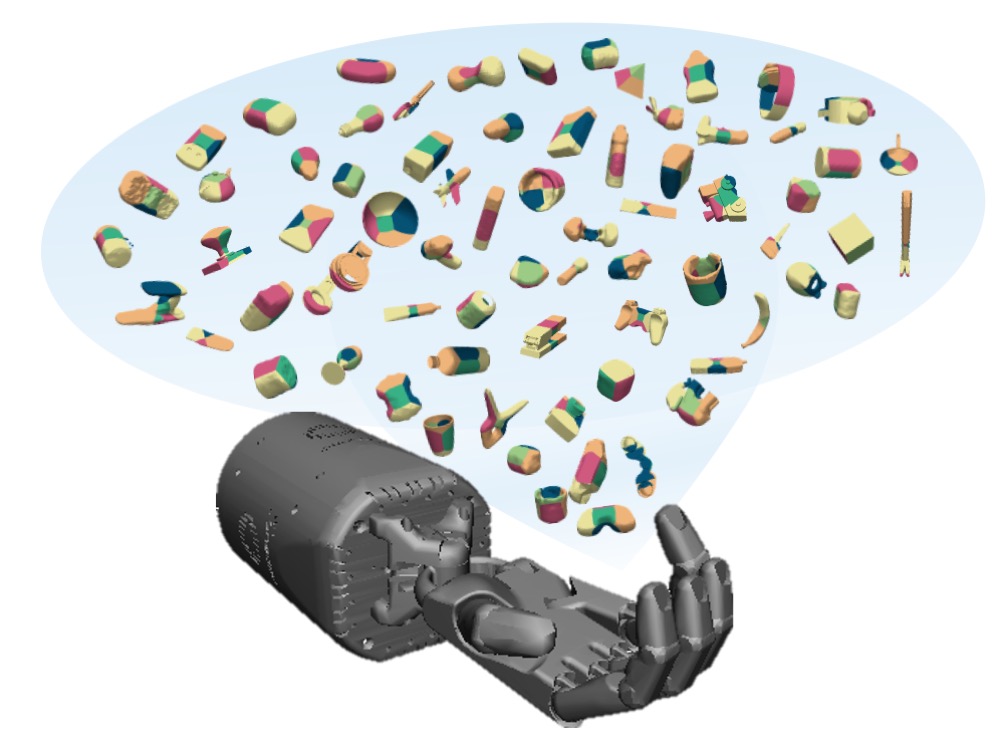}
 \vspace{-0.25in}
 \caption{\small Our goal in this work is to train a single policy that can perform in-hand manipulation on a large number of objects. We show surprising results that simple multi-task learning combined with appropriate representation not only achieves the aforementioned goal but also outperforms the single-task oracles, on both training and unseen objects.}
 \vspace{-0.1in}
 \label{fig:teaser}
\end{figure}

\begin{figure*}[t]
 \centering
 \includegraphics[width=\linewidth]{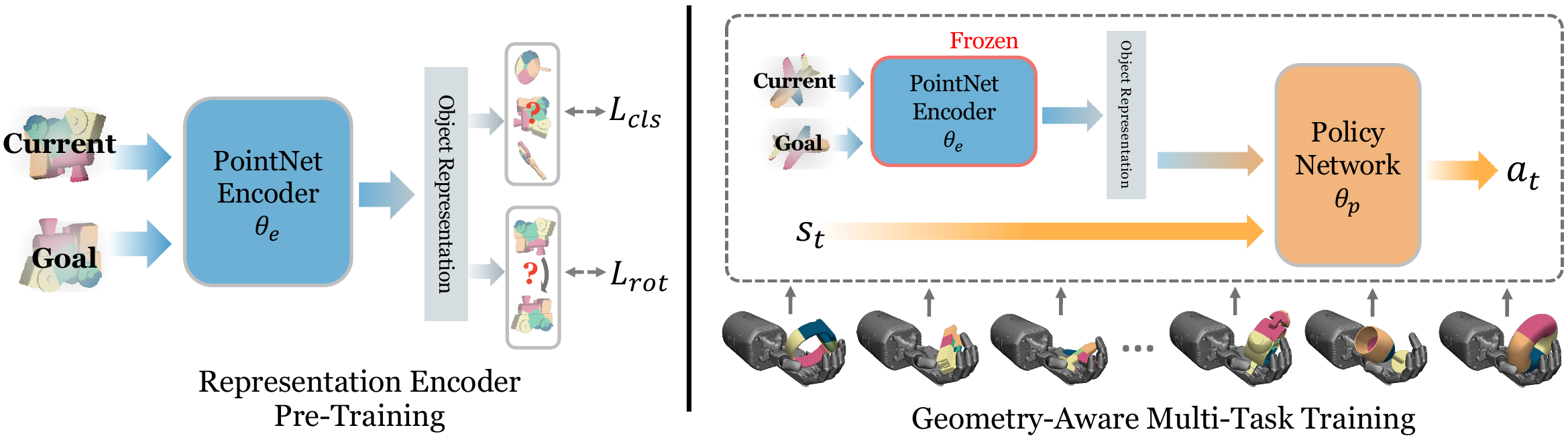}
 \vspace{-0.1in}
 \caption{\small We show that simple extensions to existing RL algorithms can produce geometry-aware dexterous manipulation policies that are robust to over 100 diverse objects.
 We first train an object representation encoder using object point clouds (left). Then we perform multi-task RL training on a large number of objects leveraging the encoded object representation (right).}
 \vspace{-0.15in}
 \label{fig:method}
\end{figure*}

However, the manipulation policies developed by these methods are typically only concerned with a single pre-determined object, making them impractical for real-world use. As there are countless objects a robot may encounter in the real world, it is infeasible to develop an individual system for each of them. Therefore, it is imperative for a policy to exhibit strong robustness to object variations, preferably in a zero-shot manner as humans do. However, generalization to completely unseen and geometrically-diverse objects for dexterous manipulation policies has been under-explored in the community, mostly due to the specious belief that such generalization is out of reach for current RL algorithms.

In this work, we show that with simple modifications, existing RL algorithms can produce strong and generalizable policies for a robotic hand with high degrees of freedom.
First, we show that contrary to the common belief of the community, a multi-task jointly-trained policy does not necessarily lead to decreased performance than if it were to be trained on each task individually. 
In fact, we show that in the context of dexterous manipulation, a multi-task policy can be a \textit{generalist} that can match the performance of those single-task \textit{specialist} policies. Second, we demonstrate that with appropriate representation that can meaningfully associate the tasks, such a multi-task policy can even outperform the single-task \textit{specialists} on completely unseen objects, where the \textit{specialists} are trained on those objects.
Furthermore, we show a linear scaling effect of better generalization when trained on more objects, revealing the possibility of future improvement. We hope that our work can serve as a key step towards building a general-purpose controller for dexterous hands. Our contributions are as follows:
\begin{itemize}
    \item We show that a multi-task policy trained on many objects can match the performance of the oracles, i.e. policies trained on those objects individually.
    \item We present a simple representation encoder that not only facilitates training but also leads to stronger zero-shot performance on unseen objects than what is possible by the oracles of those objects. 
    \item We provide detailed analysis of the potential scaling effect and the relevant design decisions of our method.
    \item We release a simulated benchmark, built upon OpenAI Gym~\cite{openaigym} and existing 3D object datasets, that contains over 100 geometrically-diverse real-world objects to encourage future studies of dexterous hand manipulation and generalization properties.
\end{itemize}

\section{Geometry-Aware Multi-Task Learning}
Prior works have shown that a standard off-policy algorithm DDPG~\cite{lillicrap2015continuous} combined with an implicit curriculum method HER~\cite{andrychowicz2017hindsight} can learn dexterous manipulation policies to control an object with simple geometries, such as a cube or an egg~\cite{plappert2018multi}.
However, whether a single policy can work well on a large number of geometrically-diverse objects has been under-explored. To this end, we first show that a simple extension to the existing algorithm can lead to a high-performing policy on diverse objects. Then we present a representation encoder module that can meaningfully leverage skills learned from different tasks, and we show the additional strong generalization properties it provides when appropriately leveraged in policy learning.

\subsection{Multi-Task Joint Training}
A straightforward way to implement a policy $\pi_{\theta_p}$, parameterized by $\theta_p$ where $p$ stands for "policy", that can manipulate a large number of $N$ objects is to train the policy on those objects jointly. We thus formulate it as a multi-task learning problem, where the goal is to optimize the sum of rewards across all $N$ objects:
\begin{align}
    \text{max}_{\theta_p}\; \mathlarger{\mathbb{E}}_{\pi_{\theta_p}} \;\mathlarger{\sum}_{i=1}^N   \mathlarger{\sum}_{t=0}^{\infty} \bigg[ \gamma^t\; r_t^i(s_{t}, \; \pi_{\theta_p}(s_t)) \bigg]
    \label{eq:policy} \\
    \text{where} \; \gamma \; \text{is the discount factor.} \nonumber
\end{align}
We refer to such a multi-task policy as \textbf{Vanilla Multi-Task Policy}. A conventional wisdom is  that a multi-task agent needs to act conservatively to simultaneously learn multiple tasks as a result of challenging joint optimization. However, to our surprise, a dexterous manipulation policy trained with this simple objective function can perform in-hand rotation remarkably well on a large number of objects. In fact, as we will show in Section~\ref{sec:joint-train}, a joint policy is comparable to individual single-task oracle policies on the trained objects while being $17$ times more sample-efficient.

We optimize the objective in Equation~\ref{eq:policy} using the same DDPG and HER setup used for training a single object, with only one modification that gradients are summed over all tasks before applying updates.

\subsection{Geometry-Aware Object Representation}
\label{sec:method-rep}
One issue arises when naively following the above setup to train a single policy on many different objects: the objects are indistinguishable from each other to the policy, and the policy would likely find a common strategy for all the objects due to the effect of joint optimization. Yet humans execute different finger gaits to manipulate geometrically different objects, and we desire the same property to emerge for a robotic hand policy. 

In fact, in addition to the standard proprioceptive inputs (i.e. joint positions and joint velocities of the robot), most existing works only consider the 6 DoF pose of the object in the state space as a result of single-object training. To explicitly model the object geometries, we propose to learn an object representation encoder based on object point clouds. A standard approach to learn useful representation such that different objects can be distinguished is to perform classification. However, as we want the representation to beneficial for the downstream in-hand rotation task and potentially useful for the policy to model hand-object contact implicitly, the representation has to be rotation-aware. A natural implementation for this might be predicting the rotation matrix based on the input point cloud. However, in the presence of many different objects, we lack a canonical coordinate frame for them, making rotation matrix prediction based on a single point cloud an ill-defined problem. We resolve this by making the encoder module take as input two copies of the same object point cloud. The first describes the current orientation at time $t$ and the second describes the desired orientation. The encoder is tasked with predicting the correct class of the object and the relative rotation matrix between the two point clouds.

We build the encoder module using a $3$-layered PointNet~\cite{qi2017pointnet} parameterized by $\theta_e$.  The training objective is defined as: 
\begin{align}
    \min_{\theta_e}\; & L_{e} = L_{cls} + \alpha L_{rot}
    \label{eq:encoder} \\
    & \text{where} \; \alpha \; \text{is weighting coefficient.} \nonumber
\end{align}

$L_{cls}$ is standard cross entropy loss for classification. Following~\cite{suwajanakorn2018discovery}, we define the rotation prediction loss $L_{rot}$ as:
\begin{align}
    & L_{rot} = 2 \; \text{arcsin} \; \bigg(\frac{1}{2 \sqrt{2}} \big\Vert \hat{R} - R \big\Vert_{F} \bigg)
    \label{eq:rotation}
\end{align}

We first pre-train the encoder on the set of training objects by randomly sampling points on object surface and randomly rotating the point cloud and its copy. After pre-training, we freeze the encoder's weights and use it in the RL training by conditioning both actor and critic networks on the encoded representation.

We refer to such multi-task policy equipped with object representation as \textbf{Geometry-Aware Multi-Task Policy}, which is the main contribution of this paper.
\section{Experiment Setup}
\label{sec:exp-setup}
We investigate our approach in the MuJoCo~\cite{todorov2012mujoco} simulated environments of the Shadow Dexterous Hand, an anthropomorphic robotic hand with $24$ degrees of freedom. We consider the task of in-hand rotation, a commonly-evaluated task for dexterous manipulation. The goal is to rotate an object to a pre-specified orientation, without dropping the object. Below we discuss the details of the evaluated objects and the environment design.

\subsection{Evaluated Objects}
\label{sec:objects}
The standard benchmarks like the Shadow Hand tasks from OpenAI Gym~\cite{openaigym} often contain a limited number of objects. To better study the generalization properties across objects, we build upon the existing environment in Gym by extending it with two object datasets, ContactDB~\cite{brahmbhatt2019contactdb} and YCB~\cite{calli2015ycb}. The two datasets contain a diverse set of real-world objects, representative of those that would often be manipulated in daily life. Since many of the object meshes are too large for in-hand manipulation, we proportionally scale down each object along its shortest axis such that the object can be fit into the palm and can be touched by the fingers. After scaling, the longest axis of all objects has an average length of $0.102m$ and a range of $[0.065m, 0.130m]$, and the shortest axis has an average length of $0.057m$ and a range of $[0.010m, 0.065m]$. Finally, to ensure the diversity of the dataset, we visually inspect each object and filter out the geometrically similar ones. In the end, we obtain a set of $114$ objects. A subset of the objects are visualized in Fig~\ref{fig:teaser}, Fig~\ref{fig:method}, and Fig~\ref{fig:spectrum}.

\subsection{Train/Test Split}
\label{sec:split}
To study the generalization properties of the proposed method, we split the objects into two disjoint sets of $85$ and $29$ objects, one for training and the other one for zero-shot testing. Because the object geometries are vastly different from each other, leading to different levels of difficulties for the manipulation policies, a random split may not ensure fair evaluations. Therefore, we use the same DDPG + HER algorithm to train an oracle single-task RL policy for each object, following the setup from~\cite{plappert2018multi}. Then we split the objects according to the success rate of its oracle, ensuring that the training and held-out objects have similar difficulties on average.

\subsection{Environment Details}
\label{sec:env-details}
\paragraph{State Space}

We define the state space $s = \{s_{r}, s_{o}, g\}$. $s_{r}$ is the proprioceptive robot states which contain joint angles and joint velocities. $s_{o}$ contains the object's Cartesian coordinates, quaternion, translational velocities, and rotational velocities. For methods that use a point cloud encoder, $s_{o}$ also includes a set of $128$ points sampled from the surface of the object and their surface normal vectors. The points are re-sampled at each timestep. $g$ is a $4$-dimensional vector specifying the desired quaternion of the object.

\paragraph{Action Space}

The action is a $20$-dimensional vector containing desired absolute joint positions for the 20 non-coupled joints of Shadow Hand. The remaining 4 joints are coupled joints which do not require control input. The action frequency is $f = 25\;$Hz.

\paragraph{Reward}
At timestep $t$, the agent receives a binary reward $r_t$ of $1$ if the angle between current object orientation and goal orientation is within $0.1$ radians and $0$ otherwise.

\paragraph{Environment Initialization and Goal Selection}

Following the environment design in~\cite{plappert2018multi}, initial position of object is set to be above the palm to avoid penetration with the hand and further perturbed with a Gaussian noise sampled from $N(0, 5\mathrm{e}{-5})$. The initial and goal orientation are sampled independently and randomly about the $z$-axis for each episode. With a diverse set of objects, we find this setting to be a good middle ground for the oracle policies we consider~\cite{plappert2018multi}, not as difficult as sampling from $SO(3)$ and not too easy as sampling from a set of pre-determined goals.

\section{Results and Ablations}
\label{sec:experiments}
Instead of focusing on how a policy performs on a specific object, our goal is to evaluate its zero-shot generalization to unseen ones. To this end, except for the oracle policies that serve as references, we evaluate all methods using the train/test split described in Section~\ref{sec:split}. Videos at~\url{https://huangwl18.github.io/geometry-dex}.

We examine the effectiveness of our proposed method by asking the following three questions:
\begin{itemize}
    \item Can vanilla multi-task policy attain competitive performance on a large number of objects?
    \item Leveraging object representation, can a single geometry-aware policy interpolate its experience and outperform single-task oracles?
    \item{What are the generalization properties of a geometry-aware policy?}
\end{itemize}

For each experiment, we report the success rate at the last timestep of an episode, averaged over $3$ training runs with different random seeds.

\subsection{Effectiveness of Multi-Task Training}
\label{sec:joint-train}
In this section, our goal is to investigate the effect of joint multi-task training alone, without leveraging object representation. We refer to such policies as \textbf{Vanilla Multi-Task Policy}. It is supplied with $s_{o}$ which only contains the object's 6 DoF pose. Below we discuss its training and generalization performance.

\begin{figure}[t]
 \centering
 \includegraphics[width=\linewidth]{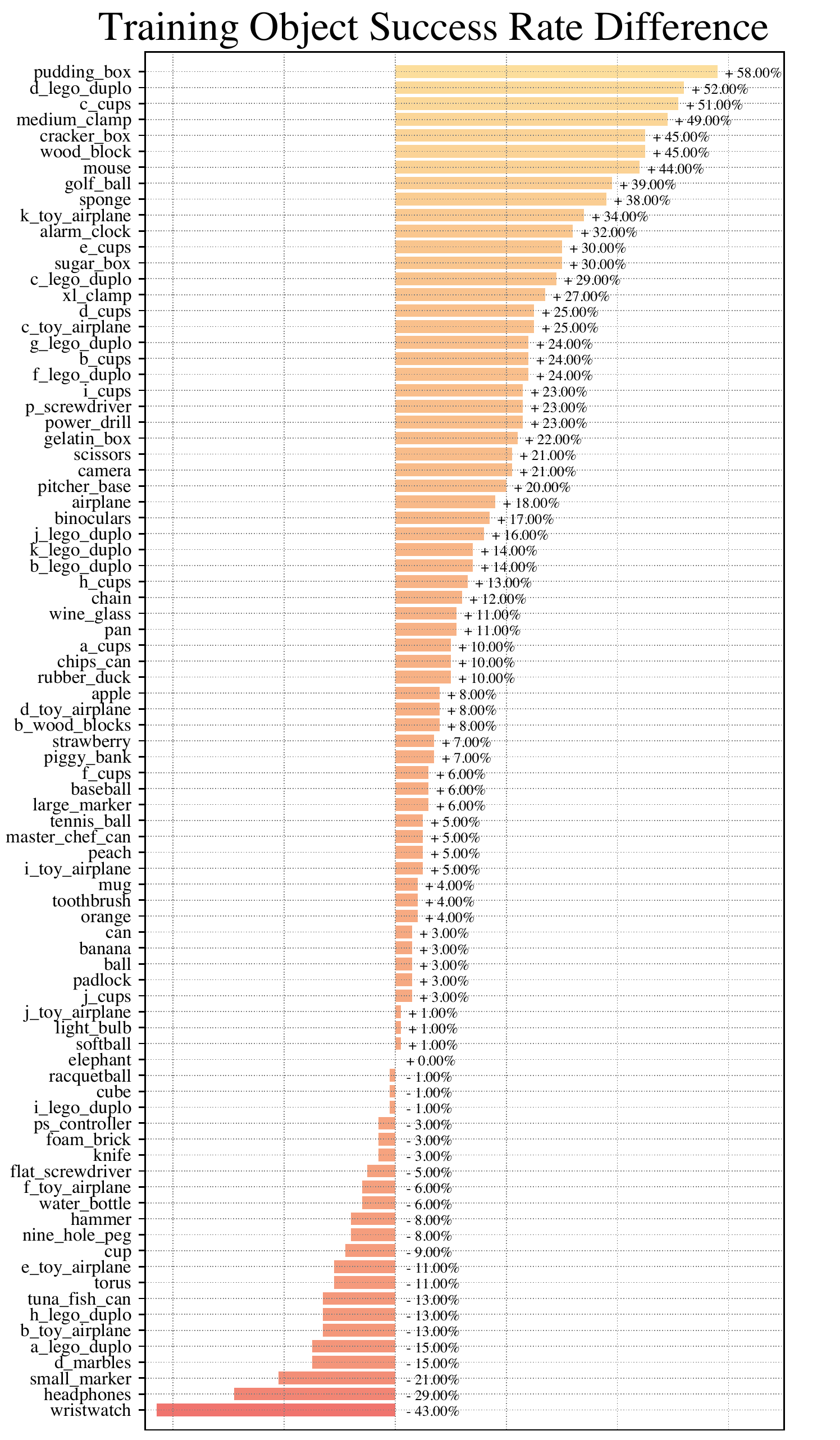}
 \caption{\small Success rate difference between geometry-aware multi-task policy and individual oracles on the 85 training objects, calculated as $\Delta S = (S_{\text{ours}} - S_{\text{oracle}})$. It shows that a single geometry-aware multi-task policy can attain even better performance than individual single-task oracles on most training objects. It demonstrates that the policy can leverage skills learned from many tasks, leading to an overall stronger policy. The success rate reported are averaged across 100 episodes.} 
 \vspace{-0.25in}
 \label{fig:multi_oracle}
\end{figure}

\paragraph{Training Performance}
We first compare the vanilla multi-task policy to the individual oracle policies trained for each object. We train the policy jointly on all 85 training objects. As shown in Fig~\ref{fig:multi_repr}, the vanilla multi-task policy is surprisingly effective in learning a joint policy for all training objects, attaining a performance on par with each object's oracle policy. In fact, it also converges to the average performance of the oracles. Furthermore, the joint-trained policy is $17$ times more sample efficient than the ensemble of single-task oracles, needing a total of $200$M samples compared to $3400$M used by all oracles.

\paragraph{Zero-Shot Generalization}
As shown in Fig~\ref{fig:multi_repr}, the vanilla multi-task policy, when evaluated on the 29 unseen objects in a zero-shot manner, performs only slightly worse than the oracles trained for the held-out objects. It shows that a simple vanilla multi-task training can produce surprisingly effective policy to handle most objects at test time. However, by ranking the objects by the performance difference between the multi-task policy and the oracles, we find that the multi-task policy generally has similar performance to the oracles on medium-sized spherical objects and performs significantly worse on irregular-shaped objects. Similar observation and analysis are also provided in Sec~\ref{sec:rep-generalization}. Overall, our suggests that the multi-task policy, though performing well on majority of objects, likely finds a common strategy that works well for majority of objects but fails to handle the uncommon ones. We refer the readers to the videos for qualitative results. 

\subsection{Role of Object Representation}
In this section, we aim to tackle geometry-awareness by investigating the role of object representation in learning multi-task dexterous manipulation policies.\footnote{We note that the representation is multi-task in nature as it is obtained by a point cloud encoder that is pre-trained on all objects, instead of a single object. Since the pre-training task is classification, a single-task representation is trivially not meaningful under this setting.} The representation is expressed as a $512$-dim feature vector taken from the last layer of a frozen point cloud encoder, described in Section~\ref{sec:method-rep}. It is updated every timestep using the latest sampled point cloud along with its rotated copy that indicates the goal orientation. We will first compare the training performance of the geometry-aware policy to its vanilla counterpart, and then we will discuss the remarkable generalization properties by making use of object representation.

\begin{figure}[t]
 \vspace{0.06in}
 \centering
 \includegraphics[scale=.48]{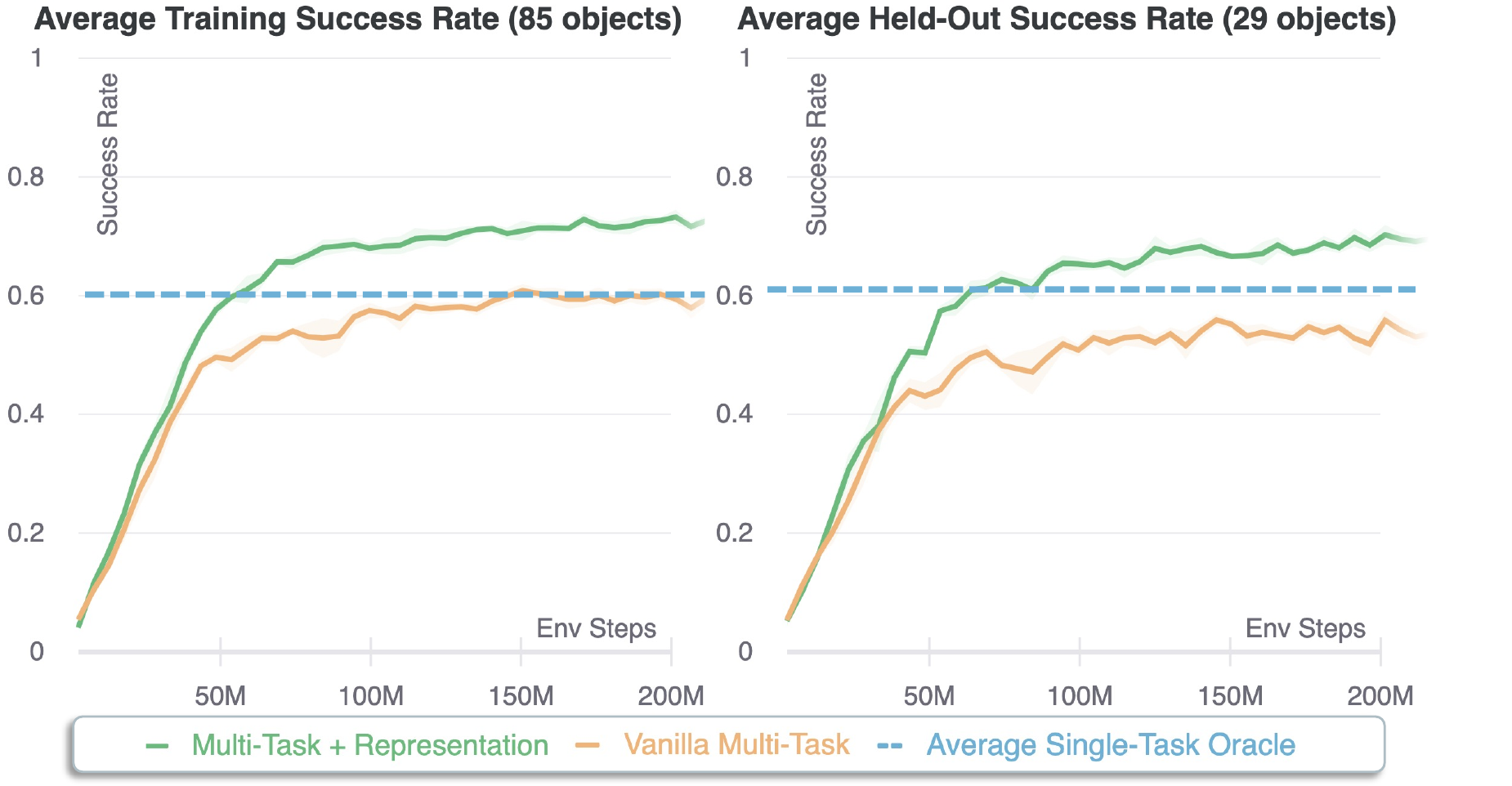}
 \caption{\small Average success rate across $85$ training objects and across $29$ held-out objects. The plot shows that multi-task joint training can lead to a surprisingly robust policy on both training and testing, with similar performance compared to the average of individual single-task oracle trained for each object. Furthermore, when combined with object representation, a joint policy can even outperform the oracles on held-out objects, in a completely zero-shot manner. The success rate reported are averaged across 425 and 145 episodes, respectively for all training objects and all held-out objects.}
 \vspace{-0.1in}
 \label{fig:multi_repr}
\end{figure}

\paragraph{Training Performance}
As shown in both Fig~\ref{fig:multi_repr}, by making use of object representation, the joint policy significantly outperforms its vanilla variant. Notably, as shown by Fig~\ref{fig:multi_oracle}, it also outperforms the single-task oracles on most objects, demonstrating that when combined with correct representation, multi-task policy does not act conservatively with decreased single-task performance to accommodate other tasks, contradictory to the common belief of the community. Furthermore, it reaches the average oracle performance at $50$M timesteps while the ensemble of oracles is trained on $3400$M timesteps, showing that a multi-task policy combined with appropriate representation can effectively interpolate and leverage its experience across all tasks and attain an overall better performance.

\begin{figure}[t]
 \centering
 \includegraphics[scale=.36]{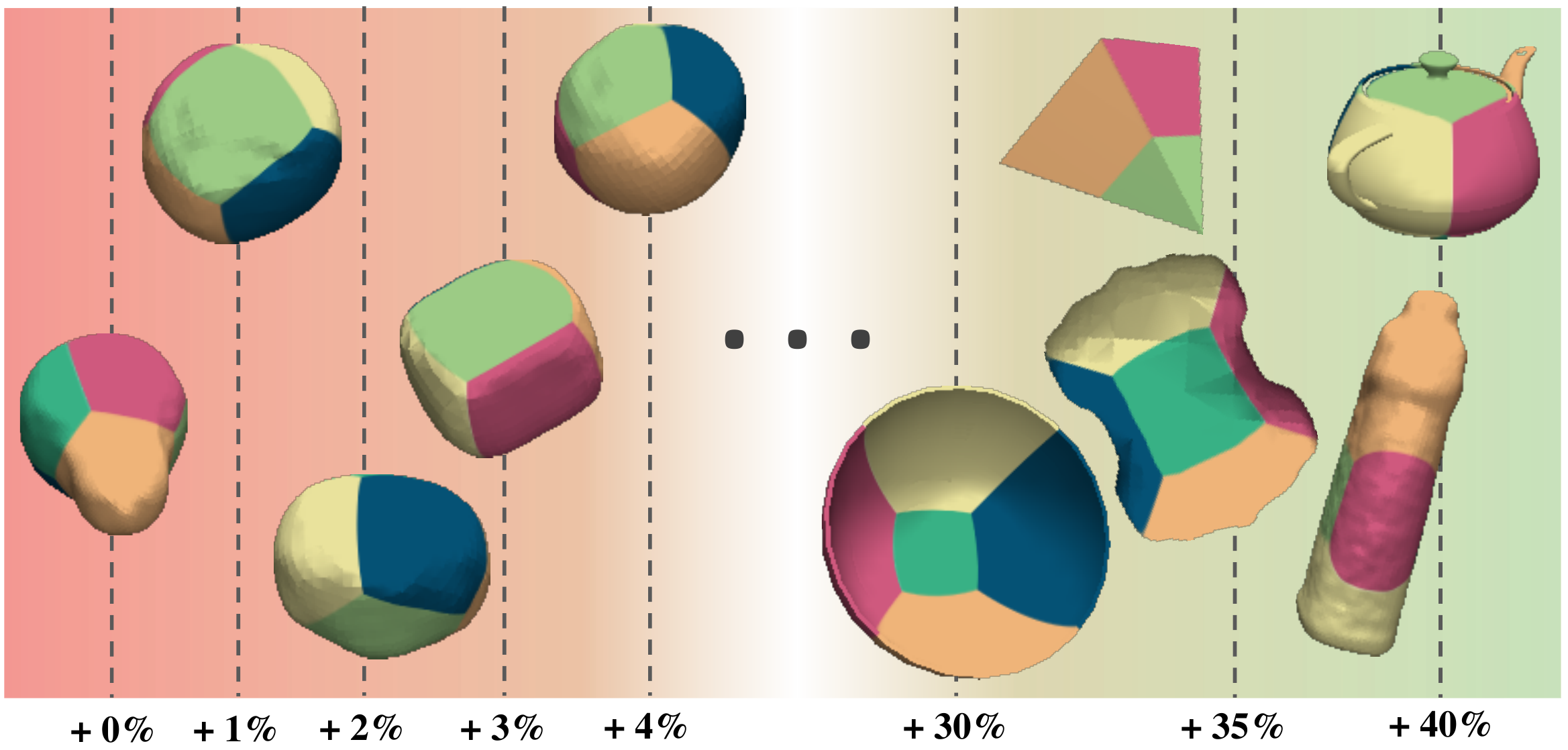}
 \caption{\small Visualization of held-out objects ranked by the performance gains of geometry-aware policy, calculated as $\Delta S = (S_{\text{ours}} - S_{\text{vanilla}})$. Notice that the gains are the highest for objects with irregular shapes and the lowest for medium-sized and spherical objects, showing the policy can effectively leverage object representation to adopt specific strategies even for challenging unseen objects.}
 \vspace{-0.23in}
 \label{fig:spectrum}
\end{figure}

\begin{figure}[t]
\vspace{0.06in}
 \centering
 \includegraphics[width=\linewidth]{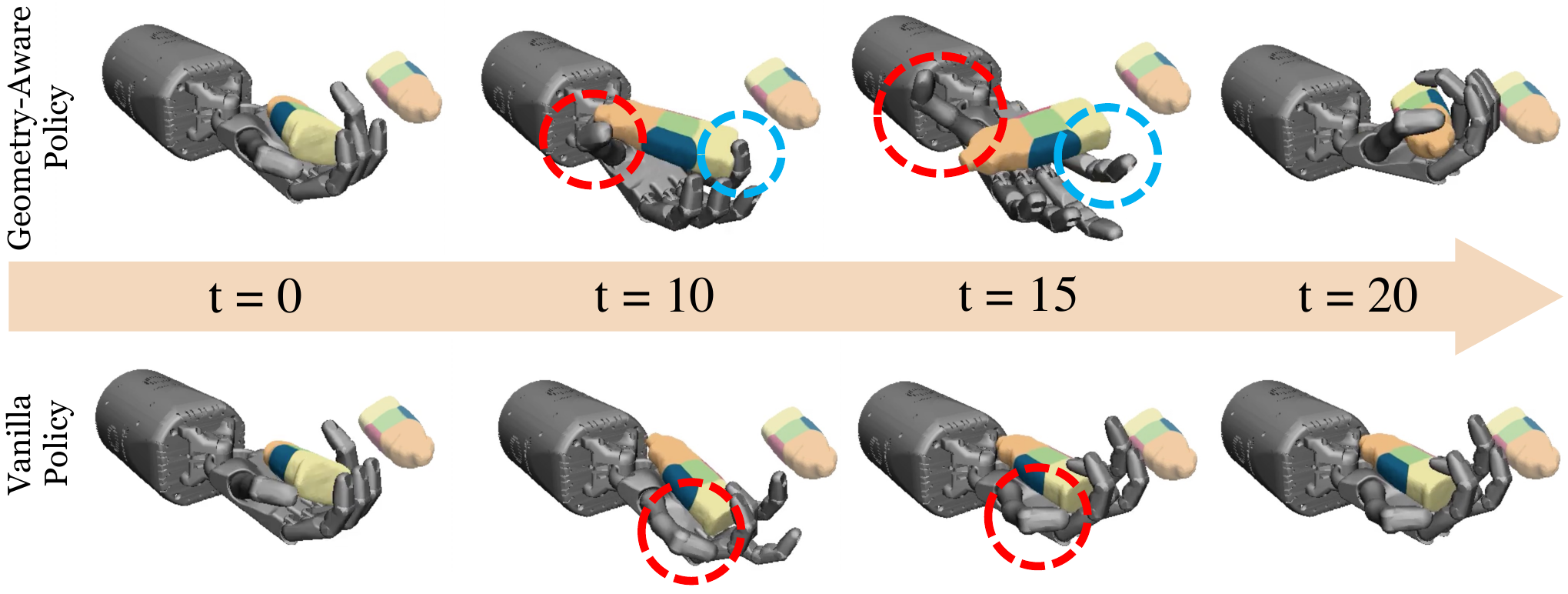}
 \caption{\small The top row shows the progression of geometry-aware policy and the bottom row shows the vanilla policy on the "bleach\_cleanser" object. Image on the right of each hand shows goal orientation. Our geometry-aware policy can reason about shape and move thumb and finger accordingly so as to rotate the object while vanilla policy can't as shown by circles.}
 \vspace{-0.1in}
 \label{fig:frames}
\end{figure}

\paragraph{Zero-Shot Generalization}
\label{sec:rep-generalization}
We evaluate our method on 29 unseen objects in a zero-shot manner. As with the comparisons for training performance, Fig~\ref{fig:multi_repr} shows that the multi-task policy trained with representation is superior than both its vanilla counterpart and even the single-task oracles that are trained on held-out objects. To validate our hypothesis that the policy is geometry-aware and can effectively handle irregular-shaped objects, we rank all 29 held-out objects by the performance gains compared to the vanilla policy and visualize 5 objects on both sides of the spectrum in Fig~\ref{fig:spectrum}, where the gains are the highest and the lowest. Not only does the policy outperforms its vanilla counterpart on all held-out objects, but notably the objects for which the performance gains are the highest have the largest variations in shapes. It shows the vanilla policy has difficulties dealing with such objects, yet the policy endowed with object representation, though never trained on these objects, learns specific strategies based on their geometries. On the other side of the spectrum, we observe that the gains are the smallest for objects that are medium-sized and spherical-shaped, suggesting that a common strategy likely works well for these objects and geometry-awareness is not needed. To further investigate the behaviors learned by our method when dealing with challenging objects, we visualize both our method and the vanilla baseline on a held-out object, shown in Fig~\ref{fig:frames}. To successfully perform a 180-degree-turn for a long and thin object, a policy must execute precise finger movement to not only grasp the object firmly but also not have the object blocked by fingers. Consistent with our hypothesis, knowing the object geometry allows the policy to precisely grip the object using the little finger at $t = 10$ and move away the thumb at $t = 15$ which was blocking the object. On the contrary, such irregular-shaped object proves to be difficult for the vanilla policy that fails to make any progress to rotate the object. We refer the readers to the video results for more qualitative comparisons.

\subsection{Ablations}

\subsubsection{Does training on more objects lead to better generalization?}

\begin{figure}[t]
 \centering
 \includegraphics[scale=.45]{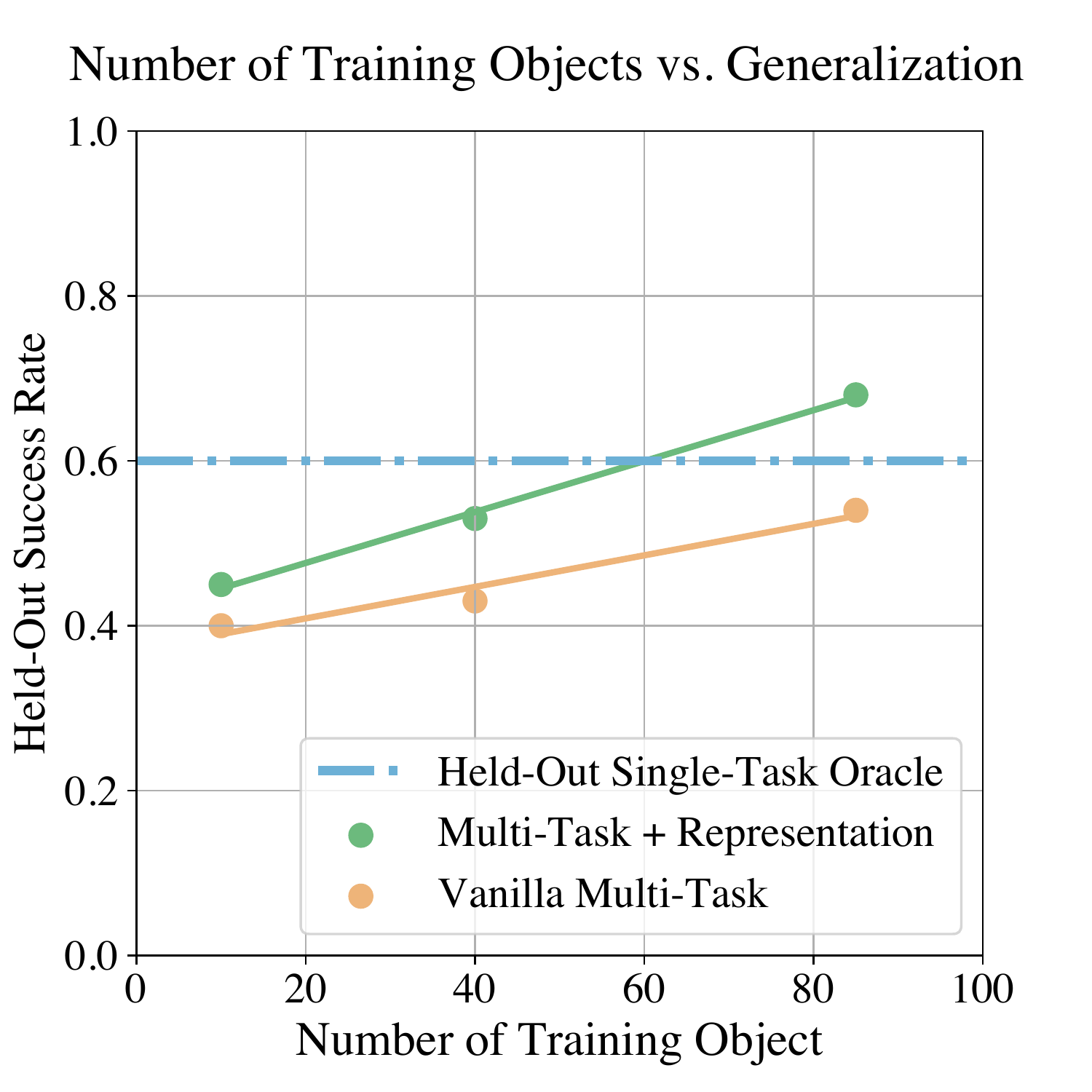}
 \vspace{-0.05in}
 \caption{\small Comparisons of the effect of the number of training objects on zero-shot generalization. More training objects would lead to a more robust policy that can even surpass single-task oracles on held-out objects.
 }
 \vspace{-0.23in}
 \label{fig:scaling}
\end{figure}

We investigate the effect of the number of training objects on the zero-shot generalization to unseen objects. For each method, we run three sets of experiments with 10, 40, and 85 training objects respectively, where each set uses 3 random seeds. Then we evaluate the trained models on the same 29 held-out objects and report the average success rate over 145 episodes. As shown in Fig~\ref{fig:scaling}, both multi-task joint training models become gradually more robust to held-out objects when trained on more objects. Notably, the method that uses object representation exhibits better scaling law than the one that doesn't. It shows a promising sign that the trend could extrapolate when trained on even more objects than what are available in this work.

\begin{table}[t]
\vspace{0.06in}
\centering
\begin{tabular}{l|l|l} %
 & w/ Object Rep. & w/o Object Rep.\\
\midrule
large\_clamp & $\bold{97.00\%}$ & $46.00\%$ \\
door\_knob & $\bold{84.00\%}$ & $65.00\%$ \\
large\_clamp & $\bold{92.00\%}$ & $74.00\%$ \\
\end{tabular}
 \caption{\small Effect of object representation used for single-task training. The objects are randomly selected from the held-out set. Even though the frozen encoder has not seen the objects in the pre-training phase, the encoded representation is still shown to be beneficial for single-task RL training, suggesting geometry-awareness is important for dexterous manipulation policies.}
\label{tab:multirep-single}
\end{table}

\subsubsection{Can object representation boost performance of single-task policies?}
Given the benefits shown for object representation when trained on a large number of objects, one natural question is what the effect is when it is used for a single task. Specifically, we take the point cloud encoder pre-trained on all 85 training objects, freeze its parameters, and use it to train an oracle policy for a held-out object from scratch, by using the same DDPG + HER algorithm. Note that the pre-trained and frozen encoder never sees the held-out object in training. Since evaluating on all 29 held-out objects are compute-intensive, we randomly select 3 objects whose average oracle success rate is similar to that of all held-out objects. We report the success rate of evaluating on 100 episodes. As shown in Table~\ref{tab:multirep-single}, the learned representation by pre-training on a large number of objects can also bring improvement to single-task training, demonstrating the object representation is universally useful for dexterous manipulation. As the majority of the current RL methods do not model object geometry explicitly, this result suggests that a geometry-aware policy may bring significant improvement to the current baseline.

\begin{table}[t]
\centering
\begin{tabular}{l|l|l} %
   & Frozen Encoder & Fine-tuned Encoder\\
\midrule
Training Success Rate & $\bold{71.88\%}$ & $61.62\%$ \\
Held-Out Success Rate & $\bold{68.80\%}$ & $59.84\%$ \\
\end{tabular}
 \caption{\small Comparisons of frozen encoder vs. fine-tuned encoder. Frozen encoder has much better performance than fine-tuned variant, whose performance is similar to that of vanilla multi-task policy without an encoder. The success rate reported are averaged across 425 and 145 episodes, respectively for 85 training objects and 29 held-out objects.}
\label{tab:frozen-finetune}
\vspace{-0.1in}
\end{table}

\subsubsection{Is it necessary to freeze the weights of point cloud encoder?}
\label{sec:frozen}
The object representation is obtained by a frozen, pre-trained point cloud encoder and has been shown to be useful in policy training. However, what if one allows the fine-tuning of the encoder? As shown in Table~\ref{tab:frozen-finetune}, a fine-tuned encoder leads to degraded performance, reducing to the same performance of a vanilla multi-task policy that doesn't have an encoder. A hypothesis is that the pre-trained encoder loses its useful representation early in the policy training due to the noisy gradient signals.

\section{Related Works}
\label{sec:related}

There has been a rapid development of anthropomorphic robotic hands~\cite{butterfass2001dlr,xu2016design}, but realizing a human-level, precise, and intelligent control of high-DoF hands has remained an unsolved problem. A series of prior works based on motion planning and optimization have been developed~\cite{sundaralingam2018geometric,bai2014dexterous,dogar2010push,kolbert2016experimental,andrews2013goal,chavan2018hand}. However, these methods require precise characterization of the systems and often fail to scale up to more complex tasks. 

Data-driven learning approaches, on the other hand, provide a promising alternative to learn behaviors directly from data. Prior works have leveraged demonstrations collected by human to either perform imitation learning or facilitate discoveries of manipulation strategies~\cite{gupta2016learning,rajeswaran2017learning,radosavovic2020state,jeong2020learning,kumar2016learning}.
Another line of works directly use reinforcement learning and learn policies without expert knowledge from humans~\cite{zhu2019dexterous,andrychowicz2020learning,akkaya2019solving,van2015learning,nagabandi2020deep,kumar2016learning,charlesworth2021solving}, but they often exhibit limited generalization capabilities:~\cite{kumar2016learning} shows generalization to different initial position of the object,~\cite{nagabandi2020deep} studies generalization to different goal settings,~\cite{rajeswaran2017learning} demonstrates generalization to the same object but with parametric variations of size and mass,~\cite{andrychowicz2020learning,akkaya2019solving} reveal the possibility of generalizing to the real world with only simulated experience but consider only one pre-determined object with simple geometries.~\cite{van2015learning,radosavovic2020state} study generalization to novel objects at test time as we do. However, they only consider generalization to one or few objects, many of which are geometrically similar to the training object.
There has been growing interest and development in multi-task learning, where a single model is trained to perform multiple tasks simultaneously~\cite{teh2017distral, yu2020gradient, hessel2019multi, espeholt2018impala}. We note that the development of better multi-task learning algorithms is orthogonal to our method. As we adopt the simplest of its form by simply summing the gradients from all tasks, one may expect further improvement by leveraging the benefits provided by the state-of-the-art multi-task learning algorithm.

Our approach of learning object representation also bears resemblance to learning task encoding in the context of multi-task reinforcement learning. This recent line of works typically encodes past experience into a task-specific embedding to interpolate learned skills from different tasks~\cite{hausman2018learning, duan2016rl, wang2016learning, mishra2017simple, rakelly2019efficient}. While these methods can be applied to dexterous manipulation to learn one policy for many objects, we propose to model the objects directly as the differences between each task are fully attributed to the differences in objects.

\noindent\textbf{Concurrent Work:} A parallel work also studies dexterous manipulation on a variety of objects~\cite{chen2021simple}. However, their approach does not condition the multi-task learning on the object geometric representation. Hence, it forces the policy to discover a common "generally" good strategy that works across many distinct objects of simpler shapes but may suffer to generalize to objects of more challenging geometries.

\section{Conclusion}
\label{sec:conclusion}
In this work, we present a simple framework for learning dexterous manipulation policies for a large variety of objects. We show that combined with simple multi-task learning and appropriate representation, existing RL algorithms can produce a single robust policy for over 100 diverse objects, which we also release as a benchmark. Overall, we hope our findings can serve as a key step towards developing general-purpose controllers for dexterous manipulation and facilitate future studies that further improve generalization across objects.

\section*{Acknowledgements}
We would like to thank Kenny Shaw and Aravind Sivakumar for feedback on the early drafts of the paper. The work was supported in part by Berkeley Deep Drive, NSF IIS-2024594 and GoodAI Research Award.

\bibliographystyle{IEEEtran}
\bibliography{IEEEabrv,main}

\end{document}